\documentclass[runningheads]{llncs}
\usepackage[T1]{fontenc}
\usepackage{graphicx}
\usepackage{amsmath}
\usepackage{nicefrac}
\usepackage{amssymb}
\usepackage[
  pdfusetitle
]{hyperref}
\usepackage{booktabs} 
\usepackage{xcolor}
\usepackage{comment}
\usepackage{courier}
\usepackage{multirow} 

\usepackage{siunitx}

\usepackage{siunitx}
\newcommand{\todo}[1]{%
  \par\noindent
  \textcolor{red}{\textbf{TODO:} #1}%
  \par
}
\newcommand{\team}[1]{\textsc{#1}}

\begin{document}
%
\title{Overview of HIPE-2026: \\
Person–Place Relation Extraction \\ from Multilingual Historical Texts}
\hypersetup{
  pdftitle={Overview of HIPE-2026: Person–Place Relation Extraction from Multilingual Historical Texts}
}

\titlerunning{CLEF-HIPE-2026}
%
\author{Juri Opitz\inst{1}\orcidID{0000-0001-6892-4574} \and
Maud Ehrmann\inst{2}\orcidID{0000-0001-9900-2193}
\and
Corina 
Raclé\inst{1}\orcidID{0009-0003-8842-8190}
\and
Andrianos Michail\inst{1}\orcidID{0009-0004-1025-7851}
\and
Matteo Romanello\inst{3,1}\orcidID{0000-0002-7406-6286}
\and
Simon  
Clematide\inst{1}\orcidID{0000-0003-1365-0662}
}
\authorrunning{J. Opitz et al.}
%
%

\institute{
   University of Zurich, Switzerland\\
    \email{}
    \and
   École Polytechnique Fédérale de Lausanne (EPFL), Lausanne, Switzerland\\
    \email{}
      \and
   Swiss Federal Institute of Technology Zurich (ETHZ), Switzerland\\
    \email{}
}

\maketitle              
\begin{abstract}
\textit{Was this person ever at that place, and if so, when?} Answering such
questions from noisy, multilingual historical documents is the central challenge of HIPE-2026, the third edition of the HIPE evaluation series. Moving from named entity recognition and linking (HIPE-2020, HIPE-2022) to reasoning about relationships between entities, HIPE-2026 targets two temporally grounded relation types: \textbf{at}, indicating that a person was present at a location at some point prior to a document's publication date, and \textbf{isAt}, indicating presence contemporaneous with that date. This paper presents the results of the evaluation campaign, which confronted 17 participating teams with the challenges of historical language variation, OCR noise, and indirect contextual cues across three languages: French, German, and English. The datasets include historical newspaper text from the nineteenth and twentieth centuries, as well as a surprise-domain generalization set drawn from early modern French literary texts. A distinctive feature of HIPE-2026 is its three-fold evaluation framework, which  assesses predictive accuracy, computational efficiency, and cross-domain generalization, reflecting the practical demands of large-scale historical document processing in the cultural heritage domain. Across more than 40 submitted runs, results reveal a wide range of strategies, from state-of-the-art large language models to lightweight task-specific classifiers, and highlight the trade-offs between accuracy, efficiency, and robustness inherent to historical relation extraction at corpus scale. System descriptions, datasets, and findings are presented and discussed, offering a detailed picture of the current state of temporally grounded relation extraction for historical documents.

\end{abstract}

\section{Introduction}
\label{sec:introduction}
Historical documents are traces of the past, recording, among other things, what people did, where they went, and how they related to one another. The large-scale digitization of historical newspapers and other cultural heritage collections has made these traces accessible at unprecedented scale, but typically yields texts that are noisy and marked by historical language variation, posing significant challenges for automatic processing.
Robust information extraction from these sources -- which span multiple languages, document types, and historical periods -- is therefore a prerequisite for many downstream uses in digital history, computational humanities, and the social sciences 
\cite{fokkens_biographynet_2014,schich_network_2014,opitz_automatic_2019,lucchini_following_2019,cardoso_construction_2020,tamper_biographysampo_2023,zhong_comprehensive_2023}. Advancing and benchmarking such methods is the driving purpose of the HIPE evaluation series\footnote{\url{https://hipe-eval.github.io/}}.

HIPE-2026\footnote{\url{https://github.com/hipe-eval/hipe-2026}} is the third edition of the series~\cite{10.1007/978-3-032-21321-1_46}. While HIPE-2020 and
HIPE-2022 focused on named entity recognition and linking in multilingual
historical documents \cite{ehrmann_overview_2020a,ehrmann_overview_2022b},
HIPE-2026 moves the series toward deeper document understanding, advancing from the identification of entities to reasoning about the relationships between them. 

The central question is whether a text evidences a relation of physical presence between a person and a place, and whether that relation holds near the publication date of the document. In short: \textit{Who was where, and
when?} Such temporally grounded person--place relations underpin the reconstruction of individuals' geographical and temporal trajectories, populating biographical knowledge graphs, supporting prosopographical research, and enabling spatial analysis of historical mobility. Advancing their extraction from historical documents is the core objective of HIPE-2026, in support of digital humanities scholarship.

To this end, the task is deliberately designed to go beyond entity co-occurrence. On one side, apparent connections may be spurious: a person and a
place mentioned in the same article may be unrelated, or linked only by affiliation rather than physical presence.
On the other, genuine connections may be hidden: relevant evidence can be indirect (e.g. a person's presence implied through an event whose location must be inferred from context), cross-sentential, or
degraded by OCR errors. Beyond the quality of evidence, the task also requires temporal reasoning: systems must distinguish between a person's presence at a place at some point in the past and their presence near the time of publication --- a distinction that is often implicit in the text.  HIPE-2026 therefore evaluates not only extraction of
explicit locative statements, but also temporally grounded
reasoning over text where evidence must be weighed rather than simply detected.

Meeting this challenge reliably and at scale requires methods and models that are fit for cultural heritage processing: accurate enough to support scholarly use, scalable enough to
process large historical collections, and versatile enough to generalize across domains and document types. For this reason, the campaign includes
three complementary profiles: an \textit{accuracy} profile on historical newspaper data,
an \textit{efficiency} profile that combines accuracy with computing resource metadata, and a
\textit{generalization} profile on a surprise French test set of literary and historiographical documents. The campaign attracted 17 participating teams, who submitted more than 40 runs across these profiles.

This lab overview defines the task (Section~\ref{sec:task}), describes the data
(Section~\ref{sec:data}), presents the evaluation framework and official metrics
(Section~\ref{sec:evaluation}), summarizes the participating systems
(Section~\ref{sec:systems}), reports the official results
(Section~\ref{sec:results}), provides additional analyses
(Section~\ref{sec:analysis}), and concludes with main lessons from the campaign
(Section~\ref{sec:conclusion}).

All shared task artifacts are publicly available: datasets can be retrieved from \url{https://github.com/hipe-eval/hipe-2026-data}, the evaluation toolkit (including submissions and results) is available at \url{https://github.com/hipe-eval/hipe-2026-eval}; Additionally, a paper with extended experiments will be released \cite{opitz_extended_2026}.

\section{Task Definition}
\label{sec:task}

HIPE-2026 frames person--place relation extraction as a document-level classification task. Given a historical document and a set of candidate (person, location) pairs extracted from it, systems must determine what kind of relationship, if any, the text supports for each candidate pair. 

For each document, systems are given access to the full text, metadata including language and publication date, and up to 16 sampled candidate pairs. 
Each candidate pair consists of one person entity and one location entity drawn from the same document. 
Person and location entities are represented by clustered mentions and may optionally carry external Wikidata identifiers. Systems must assign labels for two relation types:
\texttt{at} and \texttt{isAt}.

\begin{itemize}
    \item The \texttt{at} relation captures whether the document supports the claim that
    the person was at the place at some point before the publication date. It is a
    three-class task. The label \texttt{TRUE} indicates explicit textual evidence,
    \texttt{PROBABLE} indicates a plausible inference from contextual cues, and
    \texttt{FALSE} indicates absence of evidence or contradiction.
    \item The \texttt{isAt} relation is a narrower temporal refinement of \texttt{at}. It captures whether the person is located at the place in the immediate temporal
    horizon of the document, operationalized in the guidelines as roughly up to one
    month before publication~\cite{clef_opitz_2026}. It is a binary task with labels \texttt{TRUE} and \texttt{FALSE}.
\end{itemize}

Conceptually, \texttt{isAt=TRUE} presupposes a positive or at least plausible
\texttt{at} relation. For example, if a text reports that a person is currently
visiting Berlin, then both \texttt{at} and \texttt{isAt} should be positive. If
the same text only mentions that the person visited Berlin the previous year,
\texttt{at} may be positive while \texttt{isAt} should be \texttt{FALSE}. 
The official
scorer nevertheless evaluates both relation types independently, so that
inconsistent predictions can be measured rather than rejected.

The \texttt{PROBABLE} label is central to the task design. It reflects an
abductive view of interpretation, in which relevant meaning can be supported by
the assumptions needed to make discourse coherent rather than by explicit
surface statements alone \cite{hobbs_interpretation_1993}. In historical
person--place relation extraction, such cases include event participation,
institutional roles, travel reports, and geographically grounded narrative
contexts.


\section{Data}
\label{sec:data}

HIPE-2026 uses two evaluation domains. Domain A contains historical newspaper articles in German, English, and French, spanning the nineteenth and twentieth centuries, derived from the HIPE-2022 newspaper data, with manually created entity annotations and Wikidata links \cite{ehrmann_overview_2022b}. Domain B is a surprise test set consisting of French literature and historiographical works from the sixteenth to eighteenth centuries, included to assess out-of-domain generalization. Its entity annotations and Wikidata links come from the \textsc{FreEM}$_{\mathrm{NER}}$ corpus (version 6) \cite{gabay-etal-2022-le,gabay_2022_6481300}, which includes an extended and improved version of \textit{Presto} core \cite{blumenthal:halshs-01585010}. The \texttt{at} relationships were annotated on top of these for this shared task. Test A evaluates both \texttt{at} and \texttt{isAt}; Test B evaluates only \texttt{at}.
The competition data (version 1.0) is archived on Zenodo\footnote{\url{https://zenodo.org/records/20615690}} and released on GitHub.\footnote{\url{https://github.com/hipe-eval/HIPE-2026-data/releases/tag/v1.0}}

\begin{table}[tbp]
\centering
\caption{Overview of released training data and official test data. Label
columns give counts for \texttt{TRUE} (T), \texttt{PROBABLE} (P), and
\texttt{FALSE} (F).}
\label{tab:data-statistics}
\begin{tabular}{lcr
@{\hspace{3mm}}r@{\hspace{2mm}}r@{\hspace{3mm}}c@{\hspace{3mm}}rrrrr} 
\toprule
Split & Lang. & Docs & Pairs & Pairs/doc & Years &
\multicolumn{3}{c}{\texttt{at}} &
\multicolumn{2}{c}{\texttt{isAt}} \\
\cmidrule(lr){7-9}\cmidrule(lr){10-11}
& & & & & & T & P & F & T & F \\
\midrule
\multirow{3}{*}{Gold Train A} 
& de & 34 & 466 & 13.71 & 1818--1948 & 135 & 62 & 269 & 89 & 377 \\
& en & 35 & 307 & 8.77 & 1790--1960 & 125 & 30 & 152 & 83 & 224 \\
& fr & 35 & 478 & 13.66 & 1804--2018 & 181 & 28 & 269 & 107 & 371 \\
\midrule
\multirow{3}{*}{Silver Train A} 
& de & 64 & 893 & 13.95 & 1798--1948 & 119 & 246 & 528 & 85 & 808 \\
& en & 23 & 206 & 8.96  & 1790--1960 & 41 & 71 & 94 & 34 & 172 \\
& fr & 293 & 4110 & 14.03 & 1798--2018 & 589 & 1010 & 2511 & 429 & 3681 \\
\midrule
\multirow{3}{*}{Silver Dev A} 
& de & 32 & 432 & 13.50 & 1818--1948 & 41 & 147 & 244 & 29 & 403 \\
& en & 17 & 151 & 8.88 & 1790--1920 &29 & 54 & 68 & 18 & 133 \\
& fr & 107 & 1498 & 14.00 & 1798--1988 & 179 & 367 & 952 & 127 & 1371 \\
\midrule
\multirow{3}{*}{Gold Test A} 
& de & 19 & 238 & 12.53 & 1808--1948 & 54 & 38 & 146 & 67 & 171 \\
& en & 19 & 162 & 8.53 & 1800--1960 & 75 & 28 & 59 & 65 & 97 \\
& fr & 19 & 238 & 12.53 & 1804--1998 & 85 & 19 & 134 & 81 & 157 \\
\midrule
Gold Test B & fr & 30 & 480 & 16.00 & 1542--1797 & 130 & 60 & 290 & 0 & 480 \\
\bottomrule
\end{tabular}
\end{table}

\begin{figure}[t]
\centering

\footnotesize
\begin{quote}\footnotesize\sffamily
NOTICE. By virtue of a bil of sale issued by \textbf{Tom Scarlett, Circuit Court Clerk for Putnam County}, \textit{.ennessee}, dated on the 6tU. day of January, 1920, I will expore to sale to tne highest bidder for cash on the 31st day of January, 1920, at 1:0; o’clock, P. M. at the Courthouse door in the Town of \textit{Cookeville}, the following property, to-wit: A house and lot located in the 19th Civil District of \textit{Putnam County, Tennessee}, containing one acre, more or less, bounded on the east by the lands of \textbf{J. R. Watson}; on the north by the T. C. R. K. Co.; on the west by the lands of \textbf{Phy}; on the south by the lands of \textbf{J. R. Watson}; levied on as the lands of \textbf{D. Rittenberry}, to satisfy a Judgment in favor of \textbf{P. L. Ramsey} ag-ih st him, with interest and all the cost.i of the case. This Jan. 6th, 1920. \textbf{L. F. MILLER, Sheriff.}
\end{quote}
\setlength{\tabcolsep}{3pt}
\begin{tabular}{llcccc}
\toprule
Person & Location & \texttt{at} & \texttt{at} T/P/F & \texttt{isAt} & \texttt{isAt} T/F \\
\midrule
Tom Scarlett & Cookeville    & F    & \textbf{35.6}/\textbf{35.6}/28.9 & F & 40.0/\textbf{60.0} \\
Tom Scarlett & Putnam County & T     & \textbf{53.3}/28.9/17.8 & T  & \textbf{60.0}/40.0 \\
J. R. Watson & Tennessee     & T     & \textbf{44.4}/33.3/22.2 & T  & 22.2/\textbf{77.8} \\
D. Rittenberry & Putnam County & T   & 37.8/\textbf{40.0}/22.2 & F & 26.7/\textbf{73.3} \\
P. L. Ramsey & Tennessee     & P & 28.9/\textbf{51.1}/20.0 & T  & 26.7/\textbf{73.3} \\
L. F. Miller & Putnam County & T     & \textbf{46.7}/37.8/15.6 & T  & \textbf{53.3}/46.7 \\
\bottomrule
\end{tabular}
\caption{Illustrative example from Test A: a newspaper excerpt published in the
\textit{Putnam County Herald} on 29 January 1920, together with its
person--place relation annotations. Distribution columns report the percentages
of submitted predictions over 45 predictions: T/P/F for \texttt{at}
(\texttt{TRUE}/\texttt{PROBABLE}/\texttt{FALSE}) and T/F for \texttt{isAt}
(\texttt{TRUE}/\texttt{FALSE}); bold values indicate the majority vote.}
\label{tab:example_domainA}
\end{figure}
\begin{figure}[tp]
\centering

\begin{quote}\footnotesize\sffamily
    [French Original] Si de ce vous esmerveillez : esmerveillez vous dadvantaige de la queue des beliers de \textit{Scythie} : que pesoit plus de trente livres, \& des moutons de Surie, esquelz fault (si \textbf{Tenaud} dict vray) affuster une charrette au cul, pour la porter tant elle est longue \& pesante.  Vous ne l' avez pas telle vous aultres paillardes de plat pays.  Et [la jument] fut amenee par mer en troys carracques \& un brigantin jusques au \textit{port de Olone} en Thalmondoys Lors que \textbf{Grandgousier} la veit, Voicy (dist il) bien le cas pour porter mon filz a \textit{Paris}.  Or ca de par dieu, tout yra bien.  Il sera grand clerc on temps advenir.  Si n' estoient messieurs les bestes, nous vivrions comme clercs.  Au lendemain apres boyre (comme entendez) prindrent chemin, \textbf{Gargantua} son precepteur \textbf{Ponocrates} \& ses gens, ensemble eulx \textbf{Eudemon} le jeune paige.  Et par ce que c' estoit en temps serain \& bien attrempé, son pere luy feist faire des botes fauves.  \textbf{Babin} les nomme brodequins.  Ainsi joyeusement passerent leur grand chemin : \& tousjours grand chere : jusques au dessus de \textit{Orleans}.  
\end{quote}

\begin{quote}\footnotesize\sffamily
    [English Translation] If this astonishes you, be yet more astonished at the tails of the rams of \textit{Scythia}, which weighed more than thirty pounds, and at the sheep of Syria, to which one must (if \textbf{Tenaud} speaks true) attach a cart to their hindquarters to carry it, so long and heavy is it. You have nothing of the sort, you hussies of the flatlands. [The mare] was brought by sea in three carracks and a brigantine as far as the \textit{port of Olonne} in Talmondais. When \textbf{Grandgousier} saw it, he said: ``Here is just the thing to carry my son to \textit{Paris}. Come now, by God, all will be well. He will be a great scholar in time to come. Were it not for these gentlemen the beasts, we would live like scholars.'' The next morning, after drinking (as you will understand), they took to the road: \textbf{Gargantua}, his tutor \textbf{Ponocrates} and his men, together with \textbf{Eudemon}, the young page. And since the weather was fair and mild, his father had him fitted with tawny boots --- which \textbf{Babin} calls buskins. And so they passed merrily along their way, always in good cheer, as far as the outskirts of \textit{Orléans}.
\end{quote}
\setlength{\tabcolsep}{3pt}
\begin{tabular}{llcc}
\toprule
Person & Location & \texttt{at} & \texttt{at} T/P/F \\
\midrule
Tenaud & Scythie & F & 15.9/4.5/\textbf{79.5} \\
Grandgousier & port de Olone & P & \textbf{59.1}/13.6/27.3 \\
Gargantua & Paris & F &15.9/27.3/\textbf{56.8} \\
Ponocrates & Orleans & T & \textbf{47.7}/13.6/38.6 \\
Eudemon & Orleans & T &\textbf{47.7}/13.6/38.6 \\
Babin & Orleans & F & 27.3/27.3/\textbf{45.5} \\
\bottomrule
\end{tabular}

\caption{Illustrative example from Test B: an excerpt from Rabelais'
\textit{Gargantua and Pantagruel} (original French at the top and English
translation below) and its person--place relation annotations. The distribution
column reports the percentages of submitted \texttt{at} predictions over 44
predictions as T/P/F (\texttt{TRUE}/\texttt{PROBABLE}/\texttt{FALSE}); bold
values indicate the majority vote.}
\label{tab:example_domainB}

\end{figure}

Several preprocessing steps were applied to both domains. After conversion to the HIPE-2026 format\footnote{\url{https://github.com/hipe-eval/HIPE-2026-data/blob/main/schemas/hipe-2026-data.schema.json}}, entity mentions without a Wikidata identifier, referred to as NIL mentions, were heuristically clustered at the document level if their character similarity exceeded 75\%. Documents longer than 12,000 characters and statistical outliers in the number of locations or persons, defined as values more than three standard deviations above the mean, were then removed. From the remaining documents, person--location candidate pairs were selected from the Cartesian product, with a maximum of 16 pairs per document. Pairs were sampled with probability proportional to $\exp(-d/T)$, where $d$ is the text distance and $T=50$ controls the degree of locality, favoring textually close pairs while still allowing occasional long-range ones. 

Detailed dataset statistics are provided in Table~\ref{tab:data-statistics}. The training and validation sets of Domain A were annotated by majority vote over three runs of GPT-4.1. For each language, a subset of articles was subsequently human-annotated and released as Gold Train A data. A sample of 23 human-annotated articles was released early as illustrative examples for participants. The Domain A test set consists of 19 unreleased human-annotated articles per language. Since Domain B is used exclusively as test data, 30 documents were sampled and human-annotated for \texttt{at}, roughly matching the size of each Domain A language-specific test set. Human annotations were carried out by eight trained annotators, with at least two native annotators per language.

Table~\ref{tab:data-statistics} highlights three properties of the data. First, candidate relation extraction creates a substantial number of decisions per document: Test A contains 638 evaluated pairs across 57 documents, and Test B contains 480 evaluated pairs across 30 documents. Second, the class distributions are imbalanced: \texttt{FALSE} is the most frequent label, while \texttt{PROBABLE} is comparatively rare. Third, the surprise set is not merely another French test sample: it differs in both domain and period, and is therefore reported separately throughout the evaluation.\newline

To illustrate the annotation process, which requires fine-grained interpretation of contextual evidence, we provide an original English test set article in Figure~\ref{tab:example_domainA}. In the excerpt, person mentions are shown in bold and location mentions in italics.

The annotations in Figure \ref{tab:example_domainA} illustrate several forms of evidence.  The court clerk Tom Scarlett must be present in Putnam County shortly before the publication of the article in order to issue the bill of sale; therefore both \texttt{at} and \texttt{isAt} are set to \texttt{TRUE}. However, he may be present at any courthouse, not necessarily in Cookeville specifically, consequently both \texttt{at} and \texttt{isAt} are labeled as \texttt{FALSE}. Neighboring the estate for sale, the lands of J.R. Watson are indicated to lie within Tennessee, where he most likely also is present. It is important to note that \texttt{at} reflects the probability that an entity is located at a given place. Given a positive \texttt{at} relationship, \texttt{isAt} assesses whether the entity can be assumed to be physically present at that location within the time horizon of the article. The bill of sale is issued for D. Rittenberry's lands, implying these lands lie within Putnam County; accordingly \texttt{at} is set to \texttt{TRUE}. Being involved in lawsuits, D. Rittenberry may not necessarily be present in Putnam County, thus \texttt{isAt} is marked as \texttt{FALSE}. The judgment was ruled in favor of P.L. Ramsey, suggesting that he is likely present in Tennessee. However, he might be a creditor from outside the state, therefore \texttt{at} is set to \texttt{PROBABLE}. If he indeed is located in Tennessee, he is most likely a fellow citizen of Putnam County, consequently \texttt{isAt} is marked as \texttt{TRUE}. Note the induced complexity due to long-range pronoun resolution: As indicated by the use of first-person pronouns and the signature of L. F Miller at the end of the article, L. F Miller is the sheriff of Putnam County and assumed present there, thus both \texttt{at} and \texttt{isAt} are set to \texttt{TRUE}. Notably, ``I'' does not refer to Tom Scarlett, despite his being textually closer. 

Since Domain B does differ significantly from the newspaper data of Domain A, we also provide an example text from Test B in Figure \ref{tab:example_domainB}. Specifically, this is an excerpt from \textit{Gargantua and Pantagruel}, a series of novels by François Rabelais written in the sixteenth century. 

The annotations of the surprise text may be explained as follows: There is no evidence Tenaud ever was in Scythia, only that he described Scythian rams, therefore \texttt{at} is marked as \texttt{FALSE}. It is unknown where Grandgousier is located. It is likely that he saw this mare at the port of Olonne, but the mare also could have been brought to his location, so we set \texttt{at} to \texttt{PROBABLE} to reflect this uncertainty. Gargantua is on his way to Paris but has not arrived, therefore \texttt{at} is \texttt{FALSE} for this relationship. Gargantua and his companions Ponocrates and Eudemon are however in the lands beyond Orléans, which indicates them having passed through Orléans. Thus, we mark their relationships with Orléans as \texttt{TRUE}. Babin is only mentioned in reference to what he calls a specific type of boots, with no previous mention or any indication of being part of the travels, consequently we set \texttt{at} to \texttt{FALSE}.


\section{Evaluation Framework}
\label{sec:evaluation}

HIPE-2026 adopts three complementary evaluation profiles, accuracy, efficiency, and generalization -- each targeting a different dimension of system performance. 

\subsection{Accuracy Profile}

The official metric is macro-averaged recall, also known as balanced accuracy.
For a label $\ell$, recall is computed as
\[
    \mathrm{Recall}(\ell) =
    \frac{\#\text{correct predictions with gold label }\ell}
         {\#\text{instances with gold label }\ell}.
\]
The macro-averaged recall (MR) of a relation type is the arithmetic mean of the recalls of
its labels. This choice gives equal weight to rare and frequent labels \cite{10.1162/tacl_a_00675,sebastiani_axiomatically_2015}, which is
important for less frequent labels such as \texttt{at=PROBABLE} and \texttt{isAt=TRUE}.

For Test A, \texttt{at} is evaluated as a three-class problem and
\texttt{isAt} as a two-class problem. The language-specific profile score is
\[
    \mathrm{Score}_{A} =
    \frac{\mathrm{MacroRecall}_{\texttt{at}} +
          \mathrm{MacroRecall}_{\texttt{isAt}}}{2}.
\]
The overall Test A ranking averages this score across German, English, and
French for runs that submitted all three language files. Partial submissions are
reported only in language-specific tables. For Test B, only \texttt{at} is
evaluated, so the profile score is simply $\mathrm{MacroRecall}_{\texttt{at}}$.

\subsection{Efficiency Profile}

The efficiency profile ranks each run by combining its accuracy rank with
resource ranks derived from organizer-normalized metadata: model parameter count
and deployed model size. The official efficiency score is the mean of these
three ranks:
\[
    R_{\mathit{eff}} =
    \frac{r_{\mathit{acc}} + r_{\mathit{params}} + r_{size}}{3}.
\]
Lower efficiency scores are better. Runs with missing resource metadata are
assigned the worst resource rank.

\subsection{Generalization Profile}

The generalization profile tests system accuracy on the surprise Test B, assessing the ability of systems to transfer beyond the historical newspaper domain to literary and historiographical texts of a different period and style. Only the \texttt{at} relation is evaluated, as a three-class problem, and the profile score is $\mathrm{MacroRecall}_{\texttt{at}}$.

\subsection{Scoring and Ranking}

All profiles are scored using the evaluation toolkit available at \url{https://github.com/hipe-eval/hipe-2026-eval}. The toolkit validates submissions against the HIPE-2026 JSON schema and produces per-run diagnostic reports alongside the official scores. Systems are ranked within each profile independently; a run must cover all three languages of Test A to appear in the overall accuracy and efficiency rankings.

\section{System Descriptions}
\label{sec:systems}

Seventeen teams submitted at least one official run, alongside two
organizer-provided baselines. The evaluation repository contains 185 prediction files in total. Most teams submitted complete Test A runs covering all three languages as well as Test B runs for the surprise set; one team submitted only an English Test A run.

\subsection{Organizer Baselines}

\paragraph{\textup{\team{HIPE-Random-Baseline}}} 
This baseline assigns labels by uniform random sampling from the label inventory
of each relation type: \texttt{TRUE}, \texttt{PROBABLE}, and \texttt{FALSE} for
\texttt{at}, and \texttt{TRUE} and \texttt{FALSE} for \texttt{isAt}. It does not
use document text, metadata, entity information, or training data.

\paragraph{\textup{\team{HIPE-Ministral-Baseline}}}
This baseline is a minimal zero-shot system for person--place relation
qualification, designed as a simple and reproducible reference point for the
shared task. It uses Ministral-3-3B-Instruct-2512\footnote{\url{https://huggingface.co/mistralai/Ministral-3-3B-Instruct-2512}},
a 3-billion-parameter instruction-following language model, executed locally in
a GGUF setup with greedy decoding, i.e. temperature set to 0.0
\cite{liu2026ministral3}. The system formulates relation prediction as a
prompted classification problem using a single instruction template. For each
pre-identified person--location pair, the model is asked to assign labels for
the two target relations, \texttt{at} and \texttt{isAt}, without task-specific
fine-tuning or few-shot examples. The system operates in a straightforward
document-level loop over candidate pairs and produces structured JSON outputs.
Its design intentionally avoids retrieval augmentation, external knowledge
sources, and ensemble strategies, focusing instead on the behavior of a single
generative model under deterministic decoding. The implementation is modular,
allowing easy modification of the prompt, decoding settings, and inference
routine. Overall, the baseline is intended as a transparent reference point that
emphasizes simplicity and reproducibility, while providing a clean starting
point for exploring more advanced prompting strategies, model adaptation
techniques, or hybrid systems.

\begin{table}[t]
\centering
\scriptsize
\caption{Participating teams and coarse methodological characteristics of their
official submissions. A check mark indicates that at least one submitted run
from the team used the corresponding component. \emph{Supervised} includes both
classical supervised methods and task-specific adaptation of pretrained models,
including fine-tuning, parameter-efficient fine-tuning, and distillation. We use ``--'' to indicate missing information.}
\label{tab:teams}
\begin{tabular}{llrccccc}
\toprule
Team & Country & Runs & LLM & Superv. & PLM enc. & Feat./rules & Ext. knowl. \\
\midrule
\team{{Awakened}}        & RO & 3 & \checkmark & \checkmark & \checkmark & \checkmark & \checkmark \\
\team{{BIU\_NLP}}        & IL & 3 & \checkmark &            &            &            &            \\
\team{{DS@GT}}           & US & 3 &            & \checkmark &            & \checkmark & \checkmark \\
\team{{FI-CODE}}         & DE & 3 & \checkmark & \checkmark & \checkmark & \checkmark &            \\
\team{{FourBytes}}       & IN & 1 &  --          & --           & --           & --           &  --          \\
\team{{GippLab}}         & DE & 3 &            & \checkmark &            &            &            \\
\team{{Hansel\&Gretel}}  & IN & 3 & \checkmark & \checkmark &            & \checkmark & \checkmark \\
\team{{INSA Lyon}}       & FR & 3 & \checkmark & \checkmark & \checkmark & \checkmark &            \\
\team{{MaxFo-Ajie}}      & CN & 3 & \checkmark &            &            & \checkmark &            \\
\team{{MILRIT}}          & FR & 3 &            & \checkmark & \checkmark & \checkmark &            \\
\team{{ROSTI}}           & FR & 3 &            & \checkmark &            & \checkmark &            \\
\team{{Rittik\&Souvik}}  & IN & 1 &            & \checkmark & \checkmark &            &            \\
\team{{Spinfo}}          & DE & 3 & \checkmark &            &            & \checkmark &            \\
\team{{UMUTEAM}}         & ES & 3 & \checkmark & \checkmark & \checkmark & \checkmark &            \\
\team{{VerbaNexAI I}}    & CO & 2 &            & \checkmark & \checkmark & \checkmark &            \\
\team{{VerbaNexAI II}}   & CO & 3 & \checkmark & \checkmark & \checkmark &            &            \\
\team{{whereami}}        & EG & 2 & \checkmark & \checkmark &            & \checkmark &            \\
\midrule
\team{HIPE-Random-Baseline}     & CH & 1 &            &            &            &            &            \\
\team{HIPE-Ministral-Baseline}    & CH & 1 & \checkmark &            &            &            &            \\
\bottomrule
\end{tabular}
\end{table}

\subsection{Participating Systems}

\paragraph{{\team{{Awakened.}}}}
The team explored both prompting-based and supervised approaches for multilingual person--place relation extraction~\cite{vasile_awakened_2026}. Two submitted runs relied on Claude Sonnet 4 with carefully designed prompts and few-shot examples, while incorporating external knowledge from Wikidata, including biographical and geographical information associated with linked entities. Additional post-processing rules enforced temporal consistency using birth and death dates. A third run employed a fine-tuned XLM-RoBERTa-large model augmented with knowledge-graph-derived and pattern-based features. The supervised model further incorporated a consistency-aware objective to discourage logically incompatible label assignments. The submitted runs therefore compared knowledge-enhanced prompting with feature-augmented multilingual encoder models.

\paragraph{{\team{{BIU\_NLP.}}}}
The team approached the task as a zero- and few-shot prompting problem without task-specific fine-tuning~\cite{keinan_where_2026}. The team evaluated several open-weight instruction-following language models, including Gemma and Mistral variants. For each person--location pair, the models received a structured prompt requiring relation classification together with a short explanatory rationale. Experiments explored prompts written either in English or in the source document language, as well as different numbers of in-context examples. The submitted runs corresponded to different language model backbones and prompting configurations selected on development data.

\paragraph{{\team{{DS@GT.}}}}
DS@GT proposed a lightweight relation extraction pipeline based on dependency graphs and engineered features~\cite{wesley_lightweight_2026}. Documents were first parsed with Stanza and transformed into document-level graphs enriched with cross-sentence connections derived from entity linking, lexical similarity, and geographic containment information. From these structures, the system extracted a set of proximity-based, syntactic, and graph-derived features. Classification was then performed using either traditional machine learning ensembles or graph neural network architectures. Different combinations of graph-based and feature-based classifiers were selected for each language, enabling comparison between neural and non-neural prediction strategies while avoiding the use of pretrained language models during classification.

\paragraph{{\team{{FI-CODE.}}}}
FI-CODE investigated three diverse approaches with a focus on computational efficiency and compact model size~\cite{griesbeck_fi-code_2026}. One system (run 2) employed a rule-based strategy that inferred relations from the textual distance between person and location mentions. Another system (run 1) adapted the ITER relation extraction architecture to a relation classification setting by providing entity annotations directly as input. The third system (run 3) explored prompt-based inference with large language models, comparing alternative prompting styles, output formats, and model scales. Together, the submissions examined the trade-offs between heuristic methods, encoder-based models, and instruction-following language models.

\paragraph{{\team{{FourBytes.}}}} The team did not provide detailed information about their system. The reported efficiency metadata, including parameter count and disk size, is consistent with the XLM-RoBERTa base model \cite{conneau-etal-2020-unsupervised}, suggesting that the system may have used this model or a closely related architecture.

\paragraph{{\team{{GippLab.}}}}
GippLab explored parameter-efficient fine-tuning of multilingual LLMs from the Qwen family~\cite{yazdani_gipplab_2026,yang2025qwen3}. Multiple model sizes were adapted using LoRA, with all linear layers receiving trainable adapters. Rather than training separate systems for individual languages, the team adopted a unified multilingual setup that jointly leveraged English, French, and German training data. The submitted runs corresponded to different Qwen3.5 model sizes selected from a broader set of experiments. This design emphasized scalable multilingual adaptation through lightweight fine-tuning.

\paragraph{{\team{{Hansel\&Gretel.}}}}
Hansel\&Gretel investigated three complementary approaches spanning supervised fine-tuning, multi-agent reasoning, and prompted inference~\cite{shekhawat_frozen_2026}. The first run fine-tuned a Qwen2.5 \cite{yang2024qwen2} instruction model using supervised generation of structured JSON outputs, complemented by Wikidata-based temporal consistency rules. The second run introduced a multi-agent framework in which specialized historian and geographer models produced independent assessments that were reconciled by an arbiter model. The third run relied on chain-of-thought prompting with a large reasoning model and dynamically retrieved Wikidata information. Across all runs, external biographical and geographic knowledge was incorporated to support temporal and spatial reasoning.

\paragraph{{\team{{INSA Lyon.}}}}
The INSA Lyon system combined handcrafted features, multilingual encoders, and large language models within a unified ensemble framework~\cite{nguyen_insa_2026}. Candidate pairs were enriched with contextual and temporal information, including temporal expressions, tense-related cues, and lexical indicators. The resulting representations were processed by several model families, including feature-based classifiers, fine-tuned multilingual Transformer encoders, and zero-shot large language models. A dedicated decision layer aggregated the outputs through voting and stacking strategies. The submitted runs corresponded to the complete heterogeneous ensemble (run 1), a compact encoder-only baseline (run 2), and an ensemble configuration without large language model inference (run 3).

\paragraph{{\team{{MaxFo-Ajie.}}}}
MaxFo-Ajie developed a prompt-based multilingual relation extraction pipeline using the GPT-5.5 model accessed through an OpenAI-compatible interface~\cite{cao_few-shot_2026}. The system classified person--location pairs with structured prompts and language-matched few-shot examples drawn from the training data. The prompting strategy encouraged conservative predictions in cases of limited evidence and enforced a strict JSON output format. The submitted runs compared inline (run 1) and multi-turn document-level few-shot prompting (run 2), as well as pair-level inference (run 3). Additional components handled normalization, robust output parsing, and schema-compliant prediction generation.

\paragraph{{\team{{MILRIT.}}}}
MILRIT submitted three systems representing different approaches to relation extraction~\cite{pham_milrit_2026}. Two runs were based on GLiREL ~\cite{boylan-etal-2025-glirel}, a general-purpose relation extraction framework. One variant fine-tuned GLiREL directly on the shared-task data (run 1), while another used GLiREL-generated relation scores as features for lightweight downstream classifiers (run 2). The third run employed a multilingual DeBERTa encoder augmented with entity markers, publication-date information, and multi-view learning using large-language-model-generated input enrichments. The submissions therefore contrasted direct adaptation of a generalist relation extraction model with task-specific multilingual encoder architectures.

\paragraph{{\team{{ROSTI.}}}}
ROSTI pursued a resource-efficient approach centered on logistic regression and sparse textual representations~\cite{checchin_some_2026}. Sentences containing candidate entities were encoded using TF--IDF features derived from both sentence content and entity mentions. Separate classifiers were trained for the two target relations and for each language. Predictions were generated at the sentence-pair level and subsequently aggregated across all occurrences of a given person--location pair. The resulting system avoided external resources and large pretrained models while providing a compact baseline for multilingual historical relation extraction.

\paragraph{\textup{\team{{Rittik\&Souvik.}}}}
Rittik\&Souvik addressed the English-language setting using a transfer-learning approach based on BERT~\cite{ram_surgical_2026}. Person--location classification was formulated as a sequence-pair task through a question-answering-inspired input representation. To reduce overfitting in the low-resource setting, the team adopted a parameter-efficient fine-tuning strategy that updated only the upper layers of the encoder and the classification head. Class imbalance was addressed through cost-sensitive learning, with increased emphasis on minority labels during optimization. The system therefore combined lightweight adaptation with task-specific input reformulation.

\paragraph{\textup{\team{{Spinfo}.}}}
The team explored prompting strategies for open-weight large language models~\cite{dick_scaffolding_2026}. After evaluating several model families, the team selected a large reasoning model for all submitted runs. Person--location pairs were processed independently using structured prompts specifying task definitions, label inventories, and output constraints (run 1). Further experiments investigated the effect of temporal information extracted with HeidelTime and conditional prompting strategies that linked predictions across subtasks (run 2), and prompts translated into the document language (run 3).\footnote{We were notified by the authors about potential issues in reproduction of run 3. For more information and extended experimentation we refer the reader to the Spinfo system description paper~\cite{dick_scaffolding_2026}.} The submitted systems focused on understanding how prompting design influences multilingual historical relation extraction.

\paragraph{\textup{\team{{UMUTeam.}}}}
UMUTeam compared supervised multilingual classification and zero-shot large language model reasoning~\cite{gomez-navalon_umuteam_2026}. The supervised approach (run 1) used an XLM-RoBERTa cross-encoder with separate prediction heads for the two target relations. In parallel, two instruction-following language models were evaluated through structured prompting (run 2, run 3). All systems incorporated document metadata, publication dates, entity mentions, and local contextual evidence extracted around candidate pairs. The submitted runs therefore examined the relative strengths of fine-tuned multilingual encoders and prompt-based inference for historical relation extraction.

\paragraph{{\team{{VerbaNex AI I.}}}}
VerbaNex AI I proposed a multilingual multitask architecture built on XLM-RoBERTa and enhanced with temporal reasoning components~\cite{almanza-gonzalez_multitask_2026}. The system combined contextual representations with named entity recognition features, class balancing strategies, and hyperparameter optimization. During preprocessing, contextual evidence windows were extracted around candidate entities, while a lightweight Temporal Transformer introduced temporal positional information and semantic refinement mechanisms. The submitted approach was designed to capture both immediate and historical person--location associations in multilingual historical documents.

\paragraph{{\team{{VerbaNex AI II.}}}}
VerbaNex AI II explored two complementary paradigms: an NLI-based encoder architecture and a self-distilled instruction-following language model~\cite{morillo_person-place_2026}. The NLI system reformulated relation extraction as textual inference, pairing document content with hypotheses describing candidate relations. The submitted system relied on Nemotron-3-Nano-4B combined with DSPy-based prompt optimization and self-distillation. Optimized prompting programs were used to generate high-quality training examples, which subsequently served for LoRA-based adaptation of the same model. The final system operated directly on multilingual OCR text and employed a unified inference pipeline across languages.

\paragraph{{\team{{whereami.}}}}
The team developed a reasoning-oriented system based on multi-stage knowledge distillation~\cite{aboelwafa_distilledgemma_2026}. The approach began with an exploration of prompting strategies across several large language models to identify a suitable teacher model. The teacher was then adapted through supervised fine-tuning and used to generate reasoning traces over the training corpus. These traces were subsequently distilled into a smaller Gemma-based student model through response-level supervision. To improve prediction consistency, the system incorporated rule-based constraints covering entity compatibility, relation directionality, contextual triggers, and locality constraints.

\subsection{Discussion}

The submitted systems illustrate the broad methodological diversity anticipated by the task design. Several teams adopted prompting-based approaches using large language models, either through proprietary systems (e.g., Awakened, MaxFo-Ajie) or open-weight models (e.g., BIU\_NLP, Spinfo, UMUTeam, or FI-CODE). These approaches generally framed person--place relation extraction as an instruction-following or reasoning task, often relying on few-shot examples, structured output formats, and carefully engineered prompts. A second group focused on adapting open-weight generative models through supervised fine-tuning or parameter-efficient techniques such as LoRA, including GippLab, Hansel\&Gretel, whereami, and VerbaNex AI II. These systems explored the extent to which task-specific supervision could be transferred into compact multilingual language models while retaining the flexibility of generative architectures.

Another prominent line of work relied on encoder-based classification models, often built on multilingual Transformer encoders such as XLM-RoBERTa, DeBERTa, or NLI-oriented architectures. Examples include Awakened, INSA Lyon, MILRIT, Rittik\&Souvik, UMUTeam, and VerbaNex AI I, which typically formulated the task as supervised pair classification and incorporated contextual, temporal, or entity-specific representations. In parallel, several teams investigated lightweight alternatives that emphasized efficiency and interpretability, including DS@GT's graph-based feature extraction pipeline, ROSTI's TF--IDF and logistic regression framework, and FI-CODE's distance-based heuristics and compact relation classification models.

Many submissions further augmented their core models with external knowledge or rule-based components. Wikidata-derived biographical and geographical information was frequently used to support temporal reasoning and consistency checking, while several systems incorporated post-processing rules, temporal constraints, entity-type validation, or retrieval-based enrichment. Finally, a number of teams explored ensemble and hybrid architectures that combined heterogeneous model families, most notably INSA Lyon, DS@GT, Hansel\&Gretel, and MILRIT. Overall, the submissions demonstrate that the task can be addressed using a wide spectrum of techniques, ranging from highly efficient symbolic and statistical methods to large-scale reasoning models, reflecting the dual emphasis of the shared task on predictive quality and computational efficiency.

\section{Main Results}
\label{sec:results}

\subsection{Accuracy Profile on Test A}

Table~\ref{tab:accuracy-full} reports the full official Test A ranking for
complete submissions, including participant runs and organizer baselines. The official ranking is based on the mean of the three
language-specific Test A scores.

\begin{table}[tp]
\centering
\footnotesize
\caption{Full official Accuracy Profile ranking on Test A. Only runs submitted
for all three Test A languages are included. The language columns report
language-specific profile scores, i.e., the mean of \texttt{at} and
\texttt{isAt} macro recall. Avg. is the official Test A score, obtained by
averaging these three language-specific scores.}
\label{tab:accuracy-full}
\setlength{\tabcolsep}{5pt}
\resizebox{!}{0.72\textwidth}{%
\begin{tabular}{rrlrS[table-format=1.4]S[table-format=1.4]S[table-format=1.4]S[table-format=1.4]}
\toprule
Rank & T-Rank & Team & Run & {de} & {en} & {fr} & {Avg. $\uparrow$} \\
\midrule
1 & 1 & \team{{Spinfo}} & 1 & 0.7608 & 0.7279 & \textbf{0.7551} & \textbf{0.7479} \\
2 &  & \team{{Spinfo}} & 3 & \textbf{0.7710} & 0.6808 & 0.7349 & 0.7289 \\
3 & 2 & \team{{MaxFo-Ajie}} & 1 & 0.672 & \textbf{0.7493 }& 0.679 & 0.7001 \\
4 &  & \team{{Spinfo}} & 2 & 0.686 & 0.6427 & 0.7383 & 0.689 \\
5 & 3 & \team{{whereami}} & 1 & 0.7041 & 0.7023 & 0.6578 & 0.688 \\
6 &  & \team{{whereami}} & 2 & 0.7072 & 0.6992 & 0.6435 & 0.6833 \\
7 &  & \team{{MaxFo-Ajie}} & 2 & 0.6485 & 0.7142 & 0.6444 & 0.669 \\
8 & 4 & \team{{Awakened}} & 3 & 0.6746 & 0.6415 & 0.6854 & 0.6671 \\
9 &  & \team{{Awakened}} & 1 & 0.6739 & 0.5848 & 0.7163 & 0.6584 \\
10 &  & \team{{MaxFo-Ajie}} & 3 & 0.6335 & 0.7003 & 0.6295 & 0.6544 \\
11 & 5 & \team{{INSA Lyon}} & 1 & 0.6725 & 0.5985 & 0.6459 & 0.639 \\
12 & 6 & \team{{gipplab}} & 2 & 0.5941 & 0.6174 & 0.6697 & 0.6271 \\
13 & 7 & \team{{Hansel\&Gretel}} & 3 & 0.5994 & 0.6367 & 0.6303 & 0.6221 \\
14 &  & \team{{gipplab}} & 1 & 0.5957 & 0.6143 & 0.6322 & 0.6141 \\
15 & 8 & \team{{MILRIT}} & 3 & 0.5886 & 0.5881 & 0.6087 & 0.5951 \\
16 & 9 & \team{{UMUTEAM}} & 2 & 0.6305 & 0.5384 & 0.588 & 0.5856 \\
17 &  & Ministral baseline & 1 & 0.5553 & 0.5551 & 0.6349 & 0.5818 \\
18 & 10 & \team{{VerbaNexAI II}} & 3 & 0.5697 & 0.5812 & 0.5877 & 0.5795 \\
19 &  & \team{{Hansel\&Gretel}} & 2 & 0.5254 & 0.5866 & 0.6243 & 0.5788 \\
20 & 11 & \team{{BIU\_NLP}} & 2 & 0.505 & 0.5762 & 0.6529 & 0.5781 \\
21 &  & \team{{MILRIT}} & 1 & 0.5802 & 0.4808 & 0.6261 & 0.5623 \\
22 &  & \team{{Awakened}} & 2 & 0.5417 & 0.4989 & 0.6076 & 0.5494 \\
23 &  & \team{{Hansel\&Gretel}} & 1 & 0.521 & 0.5976 & 0.5189 & 0.5458 \\
24 &  & \team{{BIU\_NLP}} & 3 & 0.5451 & 0.5101 & 0.5617 & 0.539 \\
25 &  & \team{{VerbaNexAI II}} & 2 & 0.5022 & 0.5718 & 0.4822 & 0.5187 \\
26 & 12 & \team{{DS@GT\_HIPE}} & 1 & 0.516 & 0.5103 & 0.5162 & 0.5142 \\
27 &  & \team{{gipplab}} & 3 & 0.5181 & 0.5221 & 0.4805 & 0.5069 \\
28 &  & \team{{VerbaNexAI II}} & 1 & 0.4808 & 0.5607 & 0.4597 & 0.5004 \\
29 & 13 & \team{{VerbaNexAI I}} & 2 & 0.5071 & 0.4808 & 0.4648 & 0.4842 \\
30 &  & \team{{DS@GT\_HIPE}} & 2 & 0.4937 & 0.4401 & 0.5171 & 0.4836 \\
31 &  & \team{{DS@GT\_HIPE}} & 3 & 0.5054 & 0.4834 & 0.4424 & 0.4771 \\
32 & 14 & \team{{FI-CODE}} & 2 & 0.4678 & 0.4624 & 0.4902 & 0.4734 \\
33 &  & \team{{INSA Lyon}} & 3 & 0.4351 & 0.5107 & 0.4734 & 0.4731 \\
34 &  & \team{{INSA Lyon}} & 2 & 0.4435 & 0.5128 & 0.4561 & 0.4708 \\
35 &  & \team{{FI-CODE}} & 3 & 0.4378 & 0.4466 & 0.5091 & 0.4645 \\
36 &  & \team{{VerbaNexAI I}} & 1 & 0.4488 & 0.4757 & 0.464 & 0.4628 \\
37 & 15 & \team{{ROSTI}} & 3 & 0.5222 & 0.4541 & 0.393 & 0.4564 \\
38 &  & \team{{ROSTI}} & 2 & 0.5192 & 0.4431 & 0.3896 & 0.4507 \\
39 &  & \team{{UMUTEAM}} & 3 & 0.4527 & 0.4312 & 0.4645 & 0.4495 \\
40 &  & \team{{ROSTI}} & 1 & 0.5196 & 0.4335 & 0.3849 & 0.446 \\
41 &  & \team{{BIU\_NLP}} & 1 & 0.4495 & 0.4583 & 0.4208 & 0.4429 \\
42 &  & \team{{UMUTEAM}} & 1 & 0.4167 & 0.4702 & 0.4357 & 0.4408 \\
43 &  & \team{{FI-CODE}} & 1 & 0.4162 & 0.4313 & 0.4333 & 0.427 \\
44 &  & \team{{MILRIT}} & 2 & 0.4365 & 0.4353 & 0.4073 & 0.4264 \\
45 & 16 & \team{{FourBytes}} & 1 & 0.372 & 0.4187 & 0.4275 & 0.4061 \\
46 &  & Random baseline & 1 & 0.4058 & 0.3733 & 0.4355 & 0.4049 \\
\bottomrule
\end{tabular}
} 
\end{table}

\team{{Spinfo}} achieved the highest Test A score with run 1
(0.748), followed by \team{{MaxFo-Ajie}} run 1 (0.700) and
\team{{whereami}} run 1 (0.688). The leading run improves over the
Ministral baseline by 0.166 absolute points. The language columns also
show that no single team is uniformly dominant in every setting:
\team{{Spinfo}} leads German and French, while
\team{{MaxFo-Ajie}} leads English.

\begin{table}[t]
\centering
\footnotesize
\caption{Best language-specific Test A runs.}
\label{tab:language-winners}
\setlength{\tabcolsep}{3pt}
\begin{tabular}{llrrrr}
\toprule
Lang. & Best run & Score & \texttt{at} MR & \texttt{isAt} MR & Runner-up \\
\midrule
de & \team{{Spinfo}} run 3 & 0.771 & 0.703 & 0.839 &
\team{{Spinfo}} run 1, 0.761 \\
en & \team{{MaxFo-Ajie}} run 1 & 0.749 & 0.742 & 0.756 &
\team{{Spinfo}} run 1, 0.728 \\
fr & \team{{Spinfo}} run 1 & 0.755 & 0.679 & 0.832 &
\team{{Spinfo}} run 2, 0.738 \\
\bottomrule
\end{tabular}
\end{table}

The per-language winners in Table~\ref{tab:language-winners} suggest two
additional patterns. First, high \texttt{isAt} macro recall was often crucial
for the best German and French scores. Second, English was the only Test A
language where the top run had similar \texttt{at} and \texttt{isAt} macro
Recall, indicating a more balanced performance between the two relation types.

\subsection{Generalization Profile on Test B}

Table~\ref{tab:generalization-full} reports the best runs on the surprise
French literary test set. This profile evaluates only the \texttt{at} relation
and is intentionally separate from Test A, because it changes both domain and
time period.

\begin{table}[tp]
\centering
\footnotesize
\setlength{\tabcolsep}{3pt}
\caption{Full official Generalization Profile ranking on Test B. The official
score is \texttt{at} macro recall (MR) on the surprise-domain French test set.
Accuracy is reported as an additional diagnostic and is not used for ranking.}
\label{tab:generalization-full}
\begin{tabular}{
  r
  r
  l
  r
  S[table-format=1.4]
  S[table-format=1.4]
}
\toprule
Rank & T-Rank & Team & Run & {MR $\uparrow$} & {Acc.} \\
\midrule
1 & 1 & \team{{MaxFo-Ajie}} & 3 & 0.8163 & 0.8729 \\
2 &  & \team{{MaxFo-Ajie}} & 1 & 0.7945 & 0.8625 \\
3 &  & \team{{MaxFo-Ajie}} & 2 & 0.7712 & 0.8542 \\
4 & 2 & \team{{Spinfo}} & 3 & 0.6984 & 0.8104 \\
5 &  & \team{{Spinfo}} & 1 & 0.6910 & 0.7729 \\
6 & 3 & \team{{BIU\_NLP}} & 1 & 0.6837 & 0.8354 \\
7 &  & \team{{Spinfo}} & 2 & 0.6674 & 0.7458 \\
8 & 4 & \team{{whereami}} & 2 & 0.6665 & 0.8333 \\
9 & 5 & \team{{gipplab}} & 2 & 0.6647 & 0.7458 \\
10 & 6 & \team{{Awakened}} & 3 & 0.6613 & 0.7292 \\
11 & 7 & \team{{Hansel\&Gretel}} & 2 & 0.6349 & 0.6438 \\
12 &  & \team{{Awakened}} & 1 & 0.6338 & 0.7250 \\
13 &  & \team{{whereami}} & 1 & 0.6325 & 0.8167 \\
14 &  & \team{{Hansel\&Gretel}} & 3 & 0.6187 & 0.7708 \\
15 &  & \team{{Hansel\&Gretel}} & 1 & 0.6107 & 0.6896 \\
16 &  & \team{{gipplab}} & 1 & 0.6085 & 0.7500 \\
17 &  & \team{{BIU\_NLP}} & 3 & 0.6000 & 0.6042 \\
18 & 8 & \team{{VerbaNexAI II}} & 3 & 0.5724 & 0.7708 \\
19 & 9 & \team{{UMUTEAM}} & 2 & 0.5723 & 0.5750 \\
20 &  & \team{{Awakened}} & 2 & 0.5509 & 0.5625 \\
21 &  & \team{{gipplab}} & 3 & 0.5382 & 0.6375 \\
22 &  & \team{{BIU\_NLP}} & 2 & 0.5265 & 0.4729 \\
23 & 10 & \team{{MILRIT}} & 3 & 0.5152 & 0.5646 \\
24 &  & \team{{VerbaNexAI II}} & 2 & 0.5076 & 0.6937 \\
25 &  & Ministral baseline & 1 & 0.5062 & 0.5583 \\
26 & 11 & \team{{INSA Lyon}} & 1 & 0.4705 & 0.6604 \\
27 &  & \team{{MILRIT}} & 1 & 0.4679 & 0.6583 \\
28 &  & \team{{VerbaNexAI II}} & 1 & 0.4419 & 0.6292 \\
29 &  & \team{{INSA Lyon}} & 2 & 0.4231 & 0.6438 \\
30 &  & \team{{INSA Lyon}} & 3 & 0.3986 & 0.6375 \\
31 & 12 & \team{{DS@GT\_HIPE}} & 3 & 0.3919 & 0.5000 \\
32 & 13 & \team{{ROSTI}} & 3 & 0.3840 & 0.5104 \\
33 &  & \team{{ROSTI}} & 1 & 0.3773 & 0.5062 \\
34 & 14 & \team{{FI-CODE}} & 2 & 0.3755 & 0.3729 \\
35 &  & \team{{MILRIT}} & 2 & 0.3742 & 0.5750 \\
36 & 15 & \team{{VerbaNexAI I}} & 2 & 0.3726 & 0.4542 \\
37 &  & \team{{DS@GT\_HIPE}} & 2 & 0.3721 & 0.5000 \\
38 &  & \team{{ROSTI}} & 2 & 0.3660 & 0.4938 \\
39 &  & Random baseline & 1 & 0.3628 & 0.3604 \\
40 &  & \team{{DS@GT\_HIPE}} & 1 & 0.3626 & 0.3708 \\
41 &  & \team{{UMUTEAM}} & 3 & 0.3620 & 0.3104 \\
42 &  & \team{{FI-CODE}} & 1 & 0.3580 & 0.4437 \\
43 &  & \team{{FI-CODE}} & 3 & 0.3546 & 0.4813 \\
44 & 16 & \team{{FourBytes}} & 1 & 0.3445 & 0.2667 \\
45 &  & \team{{VerbaNexAI I}} & 1 & 0.3346 & 0.5375 \\
46 &  & \team{{UMUTEAM}} & 1 & 0.3333 & 0.6042 \\
\bottomrule
\end{tabular}
\end{table}

\team{{MaxFo-Ajie}} achieved the strongest generalization result:
its three submitted runs occupy the first three ranks in the full ranking, with
run 3 reaching 0.816 macro recall. This contrasts with Test A, where
\team{{Spinfo}} was strongest overall. The difference indicates that
in-domain newspaper performance and robustness to the literary surprise domain
should be interpreted as related but distinct capabilities.

\subsection{Efficiency Profile}

Table~\ref{tab:efficiency-full} reports the full official efficiency profile,
which combines each run's accuracy rank, parameter-count rank, and model-size
rank by their arithmetic mean.

\begin{table}[p]
\centering
\footnotesize
\setlength{\tabcolsep}{3pt}
\caption{Full official Efficiency Profile ranking on Test A. The efficiency
score is the mean of three component ranks: accuracy-profile score, parameter
count, and model size. The final column reports the official Test A
accuracy-profile score for reference. The table contains 43 runs rather than the
46 runs in the Accuracy Profile because the three MaxFo-Ajie runs opted out of
the efficiency ranking. Tied overall ranks are shown only on their first row and ordered by decreasing
accuracy-profile score.}
\label{tab:efficiency-full}
\begin{tabular}{
  rr
  l
  r
S[table-format=2.1,round-mode=places,round-precision=1]
  rrr
  S[table-format=1.4]
}
\toprule
\multicolumn{2}{c}{Ranking} & \multicolumn{2}{c}{Submission} &
{Efficiency} & \multicolumn{3}{c}{Component ranks} & {Acc.} \\
\cmidrule(lr){1-2}\cmidrule(lr){3-4}\cmidrule(lr){6-8}
Rank & T-Rank & Team Name & Run & {Eff. $\downarrow$} &
{Acc.} & {Param.} & {Size} & {Score $\uparrow$} \\
\midrule
1 & 1 & \team{{MILRIT}} & 3 & 9.67 & 12 & 8 & 9 & 0.5951 \\
2 & 2 & \team{{FI-CODE}} & 2 & 10.33 & 29 & 1 & 1 & 0.4734 \\
3 & 3 & \team{{DS@GT\_HIPE}} & 1 & 10.67 & 23 & 5 & 4 & 0.5142 \\
4 &  & \team{{DS@GT\_HIPE}} & 2 & 12.00 & 27 & 5 & 4 & 0.4836 \\
5 &  & \team{{DS@GT\_HIPE}} & 3 & 12.33 & 28 & 5 & 4 & 0.4771 \\
6 &  & \team{{MILRIT}} & 1 & 13.67 & 18 & 12 & 11 & 0.5623 \\
  & 4 & \team{{ROSTI}} & 3 & 13.67 & 34 & 4 & 3 & 0.4564 \\
  &  & \team{{ROSTI}} & 2 & 13.67 & 35 & 3 & 3 & 0.4507 \\
  &  & \team{{ROSTI}} & 1 & 13.67 & 37 & 2 & 2 & 0.4460 \\
7 &  & Random baseline & 1 & 15.00 & 43 & 1 & 1 & 0.4049 \\
8 & 5 & \team{{Awakened}} & 2 & 15.33 & 19 & 14 & 13 & 0.5494 \\
9 &  & Ministral baseline & 1 & 15.67 & 14 & 19 & 14 & 0.5818 \\
 & 6 & \team{{VerbaNexAI I}} & 2 & 15.67 & 26 & 11 & 10 & 0.4842 \\
10 & 7 & \team{{INSA Lyon}} & 2 & 16.00 & 31 & 9 & 8 & 0.4708 \\
11 & 8 & \team{{whereami}} & 1 & 16.67 & 4 & 22 & 24 & 0.6880 \\
12 &  & \team{{whereami}} & 2 & 17.00 & 5 & 22 & 24 & 0.6833 \\
   & 9 & \team{{VerbaNexAI II}} & 3 & 17.00 & 15 & 20 & 16 & 0.5795 \\
   &  & \team{{FI-CODE}} & 1 & 17.00 & 40 & 6 & 5 & 0.4270 \\
13 & 10 & \team{{UMUTEAM}} & 1 & 17.33 & 39 & 7 & 6 & 0.4408 \\
14 & 11 & \team{{gipplab}} & 2 & 17.67 & 9 & 21 & 23 & 0.6271 \\
15 &  & \team{{VerbaNexAI I}} & 1 & 18.00 & 33 & 11 & 10 & 0.4628 \\
16 &  & \team{{UMUTEAM}} & 2 & 18.33 & 13 & 20 & 22 & 0.5856 \\
17 &  & \team{{VerbaNexAI II}} & 1 & 18.67 & 25 & 16 & 15 & 0.5004 \\
18 &  & \team{{gipplab}} & 3 & 19.67 & 24 & 17 & 18 & 0.5069 \\
 & 12 & \team{{FourBytes}} & 1 & 19.67 & 42 & 10 & 7 & 0.4061 \\
19 & 13 & \team{{Hansel\&Gretel}} & 1 & 20.00 & 20 & 19 & 21 & 0.5458 \\
20 & 14 & \team{{Spinfo}} & 1 & 20.33 & 1 & 30 & 30 & 0.7479 \\
21 &  & \team{{Spinfo}} & 3 & 20.67 & 2 & 30 & 30 & 0.7289 \\
   &  & \team{{gipplab}} & 1 & 20.67 & 11 & 25 & 26 & 0.6141 \\
   &  & \team{{INSA Lyon}} & 3 & 20.67 & 30 & 15 & 17 & 0.4731 \\
22 &  & \team{{Spinfo}} & 2 & 21.00 & 3 & 30 & 30 & 0.6890 \\
23 &  & \team{{Hansel\&Gretel}} & 2 & 21.67 & 16 & 24 & 25 & 0.5788 \\
   &  & \team{{VerbaNexAI II}} & 2 & 21.67 & 22 & 23 & 20 & 0.5187 \\
24 &  & \team{{MILRIT}} & 2 & 22.00 & 41 & 13 & 12 & 0.4264 \\
25 &  & \team{{INSA Lyon}} & 1 & 22.67 & 8 & 29 & 31 & 0.6390 \\
26 &  & \team{{FI-CODE}} & 3 & 23.00 & 32 & 18 & 19 & 0.4645 \\
27 &  & \team{{Awakened}} & 3 & 23.67 & 6 & 32 & 33 & 0.6671 \\
28 &  & \team{{Awakened}} & 1 & 24.00 & 7 & 32 & 33 & 0.6584 \\
29 &  & \team{{Hansel\&Gretel}} & 3 & 24.33 & 10 & 31 & 32 & 0.6221 \\
30 & 15 & \team{{BIU\_NLP}} & 2 & 24.67 & 17 & 28 & 29 & 0.5781 \\
   &    & \team{{BIU\_NLP}} & 3 & 24.67 & 21 & 26 & 27 & 0.5390 \\
31 &  & \team{{UMUTEAM}} & 3 & 26.00 & 36 & 20 & 22 & 0.4495 \\
32 &  & \team{{BIU\_NLP}} & 1 & 31.00 & 38 & 27 & 28 & 0.4429 \\
\bottomrule
\end{tabular}
\end{table}

The efficiency profile changes the interpretation of the leaderboard.
\team{{MILRIT}} run 3 ranks first under the official efficiency metric,
despite not being the highest-accuracy run. \team{{FI-CODE}} and
\team{{DS@GT\_HIPE}} also rank highly because their resource ranks
offset lower accuracy. The result of FI-CODE is particularly informative as a baseline assessment: its heuristic classifies pairs based on textual distance. While this strategy is extremely efficient, its moderate accuracy shows that distance biases are not strong enough to solve the task, especially because the data also contains many long-distance relations between persons and places. Conversely, the most accurate Test A system, \team{{Spinfo}}, ranks lower in the official efficiency profile because
of its large reported parameter count and model size. This confirms the
usefulness of reporting efficiency separately from accuracy for large-scale
historical-text processing.

\section{Additional Analysis}
\label{sec:analysis}

\subsection{Baselines}

On Test A, the Ministral baseline scored 0.582, clearly above the random
baseline score of 0.405, but below the best participant run by a large margin.
On Test B, the same baseline scored 0.506, again above random (0.363) but far
behind the best surprise-domain run (0.816). These results make the baseline a
reasonable reference point for zero-shot LLM prompting, while showing that
task-specific system design and model size matters.

\subsection{The Role of \texttt{PROBABLE}}

The official ternary setup evaluates \texttt{PROBABLE} as a separate label for
\texttt{at}. This makes the task more demanding than binary relation detection:
systems must distinguish explicit evidence from plausible inference. Diagnostic
files show that \texttt{PROBABLE} is often the hardest label. For example, on
German Test A, \team{{Spinfo}} run 1 obtained high \texttt{FALSE}
recall (0.952) but a lower \texttt{PROBABLE} recall (0.579). The Ministral
baseline on the same file reached only 0.184 recall for \texttt{PROBABLE}.

The generated binary analysis, where \texttt{PROBABLE} is mapped to
\texttt{TRUE} for \texttt{at}, confirms that this distinction materially affects
scores. In that setting, the top Test A score rises from 0.748 to 0.842, and the
ordering of the best teams changes after the first position. We therefore treat
binary \texttt{at} results as a useful sensitivity analysis, but not as a
replacement for the official evidence-sensitive ternary task.

\subsection{System Behaviour on the Surprise Set}

Figure~\ref{tab:example_domainB} includes, alongside the gold annotations, the distribution of predictions submitted by all participating systems, offering a window into collective system behaviour on the surprise literary set. The \textbf{Tenaud}--\textit{Scythie} pair (gold: \texttt{FALSE}) was correctly handled by most systems (79.5\%), confirming that absent locative evidence for a named authority is generally well detected. More revealing are the error patterns. The \textbf{Gargantua}--\textit{Paris} pair (gold: \texttt{FALSE}) attracted 27.3\% \texttt{PROBABLE} predictions, suggesting that directional movement toward a destination is a systematic source of confusion. The \textbf{Grandgousier}--\textit{port de Olone} pair (gold: \texttt{PROBABLE}) was assigned \texttt{TRUE} by a majority of systems (59.1\%), confirming the tendency observed elsewhere to resolve locative uncertainty toward the more confident label. Finally, the \textbf{Ponocrates} and \textbf{Eudemon}--\textit{Orléans} pairs (gold: \texttt{TRUE}) show that inferring physical passage through a location from narrative context is far from trivial, with 38.6


\section{Conclusion}
\label{sec:conclusion}

HIPE-2026 extends the HIPE evaluation series from named entity recognition and linking toward document-level relation understanding in historical text. The shared task focuses on temporally grounded person--place relation extraction, requiring systems to determine not only whether a person and a location are related, but also whether the evidence supports current presence near the publication date. This formulation moves beyond entity co-occurrence and challenges systems to perform contextual, temporal, and evidence-sensitive reasoning in noisy historical documents.

The results show that the task is challenging  but tractable under current modeling approaches. On the historical newspaper benchmark, the strongest systems substantially outperformed the organizer baselines, with the best run achieving a Test A score of 0.748 compared to 0.582 for the Ministral baseline and 0.405 for the random baseline. At the same time, the leaderboard remained highly competitive, with multiple teams achieving strong performance through markedly different methodological choices. The submissions ranged from prompted large language models and parameter-efficient fine-tuning to multilingual encoder architectures, graph-based methods, feature-engineered classifiers, and lightweight rule-based systems.

Several observations emerge from the evaluation. First, language-specific performance patterns indicate that the two target relations pose different challenges. High \texttt{isAt} recall was particularly important for strong performance in German and French, whereas the best English systems achieved a more balanced trade-off between \texttt{at} and \texttt{isAt}. Second, the surprise-domain evaluation confirms that in-domain accuracy and cross-domain robustness are distinct capabilities. While \team{{Spinfo}} achieved the highest overall score on the historical newspaper benchmark, \team{{MaxFo-Ajie}} dominated the literary surprise set, demonstrating that successful transfer to substantially different genres and historical periods remains a separate challenge.

The ternary formulation of \texttt{at} proved to be an important aspect of the task design. Distinguishing between \texttt{TRUE} and \texttt{PROBABLE} requires systems to separate explicit textual evidence from plausible contextual inference, making the task substantially more demanding than binary relation detection. The binary sensitivity analysis showed that collapsing \texttt{PROBABLE} into \texttt{TRUE} increases the best Test A score from 0.748 to 0.842 and changes the ordering of several leading systems. This confirms that modeling uncertainty and abductive inference is a central challenge of historical person--place relation extraction rather than a marginal edge case.

The diversity of successful approaches is equally noteworthy. Many of the highest-ranked systems relied on large language models, either through prompting or supervised adaptation, suggesting that modern generative models are effective at integrating dispersed contextual and temporal evidence. However, competitive results were also obtained with encoder-based architectures and feature-driven approaches, indicating that strong performance does not depend on a single modeling paradigm. The task therefore provides a useful testbed for comparing reasoning-oriented and classification-oriented approaches under a common evaluation framework.

The efficiency profile further highlights the importance of considering computational cost alongside predictive quality. Systems with the highest accuracy often relied on comparatively large models, whereas methods based on compact encoders, engineered features, or lightweight classifiers achieved substantially better efficiency rankings. In particular, the strong efficiency results of MILRIT, FI-CODE, and DS@GT demonstrate that practical large-scale deployment on cultural-heritage collections may favor different solutions than those that maximize accuracy alone. Reporting both dimensions therefore provides a more complete picture of system suitability for real-world historical-text processing.

Overall, HIPE-2026 establishes a benchmark for evidence-sensitive and temporally grounded relation extraction in historical documents. The results show clear progress beyond entity recognition while also revealing persistent challenges involving temporal interpretation, implicit evidence, and domain transfer. Future work can investigate the remaining sources of error in temporal interpretation and plausible inference, extend the benchmark to additional relation types and historical genres, and explore approaches that improve cross-domain robustness without sacrificing computational efficiency.

\begin{credits}
\subsubsection{\ackname} 

The CLEF-HIPE-2026 organizing team thanks the CLEF 2026 Conference and Evaluation Labs Committees for hosting the task and Simon Gabay for helping with the annotation of the surprise test data. This work is carried out within the framework of the Impresso -- Media Monitoring of the Past project, funded by the Swiss National Science Foundation under grant No.\ CRSII5\_213585 and by the Luxembourg National Research Fund under grant No.\ 17498891.

\subsubsection{\discintname}
The authors have no competing interests to declare that are relevant to the content of this article. 
\end{credits}
%
%
%
\bibliographystyle{splncs04}
\bibliography{2026-HIPE-CLEF-Overviews,2026-NotInZotero}

\end{document}